\documentclass[conference]{IEEEtran}
\usepackage{times}

\usepackage[numbers]{natbib}
\usepackage{multicol}
\usepackage[bookmarks=true]{hyperref}

\usepackage{amsmath}

\usepackage{hyperref}       
\usepackage{url}            
\usepackage{booktabs}       
\usepackage{amsfonts}       
\usepackage{nicefrac}       
\usepackage{microtype}      
\usepackage{xcolor}         

\usepackage{threeparttable}
\usepackage{amssymb}
\usepackage{amsmath} 
\usepackage{booktabs}
\usepackage{multirow}
\usepackage{graphicx}
\usepackage{threeparttable}
\usepackage{colortbl}
\usepackage{float}
\usepackage{stfloats}
\usepackage{amssymb}
\usepackage{pdflscape}
\usepackage{gensymb}
\usepackage{authblk}

\begin{document}

\title{Spatiotemporal Contrastive Learning for Cross-View Video Localization in Unstructured Off-road Terrains}

\author{Zhiyun Deng, Dongmyeong Lee, Amanda Adkins, Jesse Quattrociocchi, Christian Ellis, Joydeep Biswas}

\maketitle

\setcounter{footnote}{0}
\renewcommand{\thefootnote}{}
\footnotetext{The authors are with The University of Texas at Austin, TX, USA. Email: \texttt{\{zdeng, domlee, amanda.adkins, jesse.quattrociocchi, christian.ellis, joydeepb\}@utexas.edu}.}
\renewcommand{\thefootnote}{\arabic{footnote}}

{\small
\begin{abstract}

Robust cross-view 3-DoF localization in GPS-denied, off-road environments remains a fundamental challenge due to two key factors: (1) perceptual ambiguities arising from repetitive vegetation and unstructured terrain that lack distinctive visual features, and (2) seasonal appearance shifts that alter scene characteristics well beyond chromatic variation, making it difficult to match current ground-view observations to outdated satellite imagery.
To address these challenges, we introduce \textbf{MoViX}, a self-supervised cross-view video localization framework that learns representations robust to viewpoint and seasonal variation, while preserving directional awareness critical for accurate localization.
MoViX employs a pose-dependent positive sampling strategy to enhance directional discrimination and temporally aligned hard negative mining to discourage shortcut learning from seasonal appearance cues.
A motion-informed frame sampler selects spatially diverse video frames, and a lightweight, quality-aware temporal aggregator prioritizes frames with strong geometric alignment while downweighting ambiguous ones.
At inference, MoViX operates within a Monte Carlo Localization framework, replacing traditional handcrafted measurement models with a learned cross-view neural matching module. Belief updates are modulated through entropy-guided temperature scaling, allowing the filter to maintain multiple pose hypotheses under visual ambiguity and converge confidently when reliable evidence is observed.
We evaluate MoViX on the TartanDrive 2.0 dataset, training on less than 30 minutes of driving data and testing over 12.29 km of trajectories. Despite using temporally mismatched satellite imagery, MoViX localizes within 25 meters of ground truth for 93\% of the time and within 50 meters for 100\% in unseen regions, outperforming state-of-the-art baselines without environment-specific tuning. We also showcase its generalization on a real-world off-road dataset collected in a geographically distinct location with a different robot platform. 
The code will be made publicly available upon publication. A demonstration video is available at: \href{https://youtu.be/y5wL8nUEuH0}{https://youtu.be/y5wL8nUEuH0}.

\end{abstract}
}

\IEEEpeerreviewmaketitle

\section{Introduction}

Cross-view visual localization plays a critical role in vision-guided navigation, autonomous driving, and augmented reality by enabling global positioning in GNSS-denied environments. This is achieved by matching ground-level observations with aerial-view reference maps, such as geo-tagged satellite imagery. While recent methods have demonstrated strong performance in structured urban environments \cite{wang2023fine,shi2023boosting}, they face major challenges in off-road settings due to two key issues: (1) perceptual ambiguity arising from repetitive, low-texture natural scenes (e.g., forest trails) that lack distinctive landmarks, and (2) substantial visual appearance changes across seasons—driven by vegetation variations in foliage density, color, and coverage—which make it difficult to reliably match current ground views with outdated satellite imagery.

In this work, we tackle the problem of cross-view video localization in GNSS-denied off-road environments using an uncalibrated monocular RGB camera. 
The goal is to estimate the vehicle's 3-Degree-of-Freedom (DoF) pose and trajectory over time by aligning ground-level video with geo-referenced satellite imagery, even if outdated or seasonally mismatched. This lightweight setup reduces system cost and complexity, while the use of an uncalibrated camera eliminates the need for camera-specific calibration procedures, simplifying deployment across different vehicles or robots equipped with front-facing cameras, regardless of variations in installation height or tilt angle.
Nonetheless, this problem remains highly challenging due to the narrow field of view, the extreme viewpoint disparity between ground and aerial images, and the limited availability of distinctive features in natural, unstructured environments.

\begin{figure}
    \centering
    \includegraphics[width=1\linewidth]{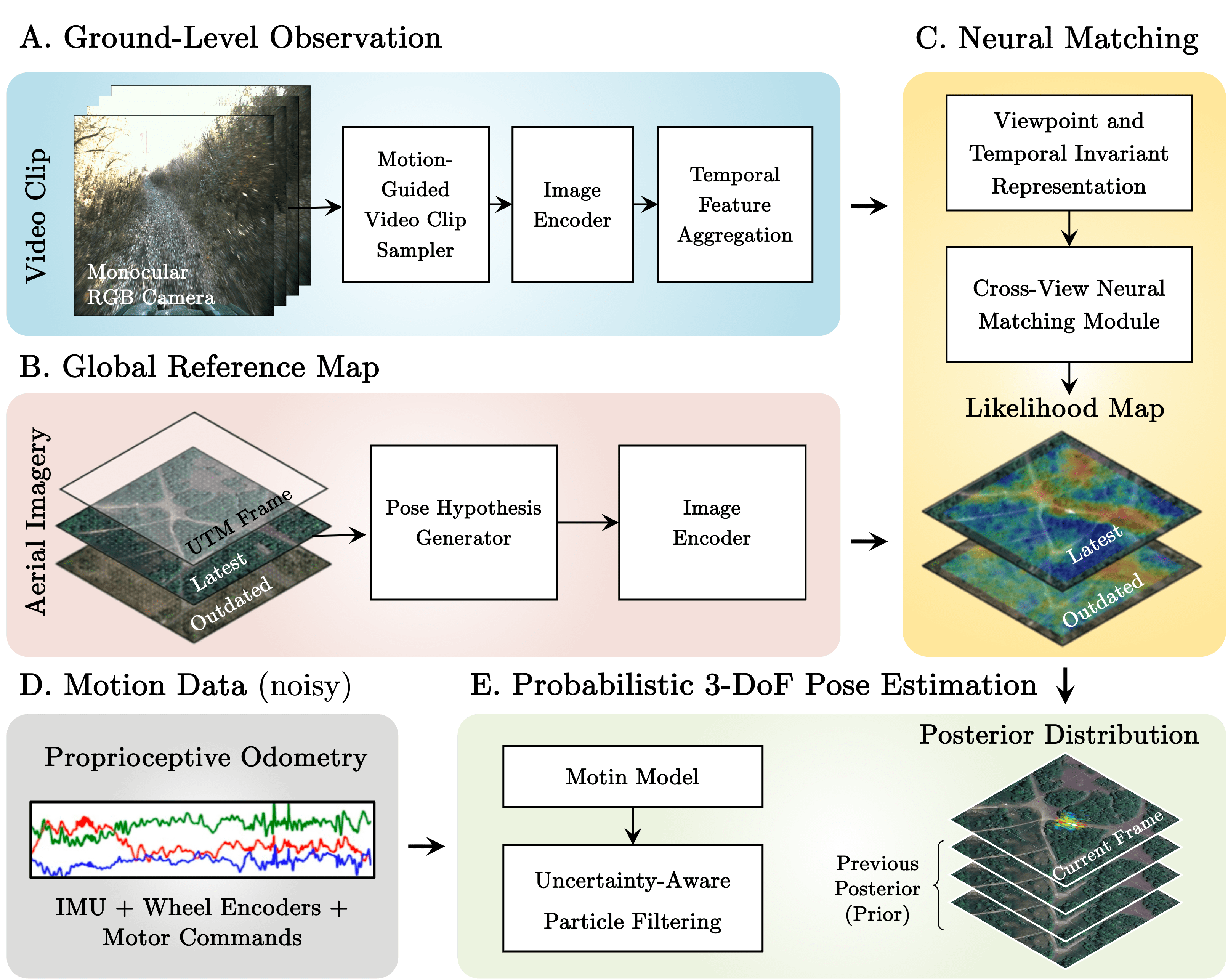}
    \vspace{-6mm}
    \caption{Overview of our proposed approach, MoViX. (A) A motion-guided video clip sampler selects informative frames from ground-level observations, which are encoded and temporally aggregated into viewpoint- and temporal-invariant, yet orientation-sensitive representations. (B) Unconstrained aerial imagery—whether outdated or seasonally mismatched—is cropped into smaller aerial patches corresponding to each pose hypothesis and then processed through a shared encoder. (C) A cross-view neural matching module compares embeddings from both branches to compute similarity scores over the hypothesized poses. (D) Noisy proprioceptive motion data serve as a dynamic prior. (E) A neural-augmented particle filter fuses visual likelihoods with motion priors to perform uncertainty-aware Bayesian inference of 3-DoF pose and trajectory estimation over time.}
    \label{fig:overview}
    \vspace{-6mm}
\end{figure}

Prior work has sought to overcome the viewpoint gap through explicit modality conversion. Some approaches synthesize bird’s-eye or overhead views from ground imagery using multi-camera systems~\cite{jin2024bevrender}, enabling subsequent template matching~\cite{klammer2024bevloc} or neural matching~\cite{sarlin2023orienternet} with aerial maps. Others take the reverse approach, generating plausible ground-level perspectives from aerial imagery~\cite{toker2021coming}. More recent methods employ contrastive learning~\cite{downes2023wide} to align disparate views without direct image conversion, while others demonstrate that aggregating video frames over time can improve localization accuracy~\cite {vyas2022gama, pillai2024garet}. Despite these advances, many of these techniques rely on complex network architectures and incur high computational costs. Moreover, their dependence on the latest satellite imagery makes them vulnerable to temporal inconsistencies.

Although data augmentation can improve robustness to illumination changes \cite{fan2024cross}, it fails to address the deeper structural transformations introduced by seasonal variation. As our experiments show, color-based augmentations are insufficient to capture long-term appearance changes caused by vegetation growth cycles and terrain evolution. We argue that addressing this challenge requires learning viewpoint- and temporally-invariant, yet orientation-sensitive representations that encode stable geometric structures across seasons. Moreover, instead of committing to potentially erroneous estimates based on single-frame observations \cite{shi2022accurate}, we advocate for a belief-space formulation that preserves multiple hypotheses in ambiguous regions and converges only when confident evidence becomes available.

To this end, we introduce \textbf{MoViX}, a \underline{Cross-view} \underline{Vi}deo localization framework with \underline{Mo}tion integration. MoViX learns the viewpoint and temporal-invariant representation through self-supervised spatiotemporal contrastive learning, using GPS-tagged ground-level observations and historical satellite imagery for training.
Our cross-view neural matching module includes a pose-dependent positive sampling strategy and temporally matched hard negative mining to encourage spatial and directional grounding. We introduce a motion-informed frame sampler that selects spatially distributed frames for robust alignment. A lightweight, quality-aware temporal aggregator is then used to emphasize informative frames and suppress ambiguous ones. At inference time, MoViX is deployed within a Monte Carlo Localization (MCL) framework as a learned observation model. By replacing handcrafted likelihoods with our perceptual model, and modulating update confidence through belief entropy (computed via kernel density estimation), the system dynamically maintains multiple pose hypotheses in ambiguous scenes and ensures confident convergence when discriminative evidence is present.

Our main contributions are summarized below:
(1) First, we propose MoViX, a self-supervised spatiotemporal contrastive learning approach for cross-view video localization that learns representations robust to viewpoint and seasonal variation, while preserving directional cues critical for accurate localization.
(2) Second, we design a motion-informed frame sampling strategy and a lightweight quality-aware temporal feature aggregation module to adapt our cross-view neural matching module to video input.
(3) Third, we integrate our perception module into a neural-augmented probabilistic localization framework with entropy-aware belief-space inference, enabling accurate and robust 3-DoF pose and trajectory estimation without environment-specific tuning.
(4) Lastly, we validate MoViX on the TartanDrive 2.0 benchmark \cite{sivaprakasam2024tartandrive} and a real-world off-road dataset, demonstrating strong generalization to out-of-distribution settings across both spatial (unseen areas) and temporal (seasonal shift) domains without retraining.

\section{Problem Formulation}

\noindent \textbf{Problem Definition.}  
We address the problem of 3-DoF global localization for ground vehicles in GNSS-denied environments by leveraging geo-tagged satellite imagery as a global reference map for cross-view visual matching. The system relies on an uncalibrated monocular camera with a limited field-of-view, complemented by proprioceptive odometry and IMU data commonly available in standard navigation stacks.

\noindent \textbf{Cross-View Neural Matching.}  
Given a sequence of ground-level RGB images $\mathcal{G}_i \in \mathbb{R}^{w \times h \times 3}$, we aim to estimate the 3-DoF pose trajectory $\hat{\mathbf{p}}_t = (\hat{x}_t, \hat{y}_t, \hat{\theta}_t)$ over time $t$, where $(\hat{x}_t, \hat{y}_t)$ represents the position in Universal Transverse Mercator (UTM) coordinates and $\hat{\theta}_t \in [-\pi, \pi]$ denotes the heading angle measured counterclockwise from the East.

At each timestep $t$, a short ground-view video clip $\mathcal{V}_{\mathcal{G}}$ is constructed from a sampled subset of the ground-level image sequence. The task is to determine the most probable vehicle pose by matching this clip against a set of geo-referenced aerial images $\{ \mathcal{A}_k \in \mathbb{R}^{w \times h \times 3} \}$, where each aerial patch corresponds to a hypothesized 3-DoF pose $\mathbf{p}^{\mathcal{A}}_k = (x_k, y_k, \theta_k)$.

To perform this matching, we employ a neural network $f_\Phi$, parameterized by weights $\Phi$, that computes a similarity score between the query ground-view clip and each candidate aerial patch:
\[
f_\Phi(\mathcal{V}_\mathcal{G}, \mathcal{A}_k) \mapsto s_k \in [0, 1],
\]
where $s_k$ represents the likelihood of the vehicle being at pose $\mathbf{p}^{\mathcal{A}}_k$. In the frame-level localization task, the final pose estimate is selected as the candidate with the highest similarity:
\begin{equation}\label{eq:p_argmax}
\hat{\mathbf{p}}_t = \mathbf{p}^{\mathcal{A}}_{k^*}, \quad \text{where} \quad k^* = \arg\max_k f_\Phi(\mathcal{V}_\mathcal{G}, \mathcal{A}_k).
\end{equation}
\noindent \textbf{Probabilistic Framework Integration.}  
To incorporate uncertainty and exploit motion priors, we embed the neural cross-view matching module within an MCL framework for probabilistic trajectory estimation, following the principles outlined in~\cite{thrun2005probabilistic}.

The belief over the vehicle pose at time $t$ is approximated by $M$ weighted particles:
\begin{equation}
\mathcal{S}_t = \left\{ \left( \mathbf{p}_t^{[n]}, w_t^{[n]} \right) \right\}_{n=1}^{M},
\end{equation}
where $\mathbf{p}_t^{[n]} = (x_t^{[n]}, y_t^{[n]}, \theta_t^{[n]})$ represents a pose hypothesis and $w_t^{[n]}$ its weight.

Particles are initialized around a prior pose with Gaussian noise and propagated using a motion model informed by control inputs with translation and rotation noise injected. On receiving a ground-view observation, the neural network evaluates the likelihood of each particle:
\begin{equation}
s_t^{[n]} = f_\Phi\left(\mathcal{V}_\mathcal{G}, \mathcal{A}\left(\mathbf{p}_t^{[n]}\right)\right),
\end{equation}
with $\mathcal{A}\left(\mathbf{p}_t^{[n]}\right)$ denoting the aerial patch at $\mathbf{p}_t^{[n]}$. Weights are updated by:
\begin{equation}
w_t^{[n]} \propto \exp\left(\lambda_t \, s_t^{[n]}\right),
\end{equation}
where $\lambda_t$ adjusts the sharpness of the measurement likelihood. The final pose estimate is computed as the weighted average of all particle hypotheses:

\begin{equation}
\label{eq:p_avg}
\hat{\mathbf{p}}_t = \sum_{n=1}^{M} w_t^{[n]} \mathbf{p}_t^{[n]}.
\end{equation}

\noindent While the frame-level localization selects the most likely pose via an argmax (Eq.~\eqref{eq:p_argmax}) over similarity scores,  particle filtering estimates it as a weighted average (Eq.~\eqref{eq:p_avg}) over all hypotheses, enabling reasoning over multiple concurrent pose beliefs.

\begin{figure*}
    \centering
    \includegraphics[width=0.9\linewidth]{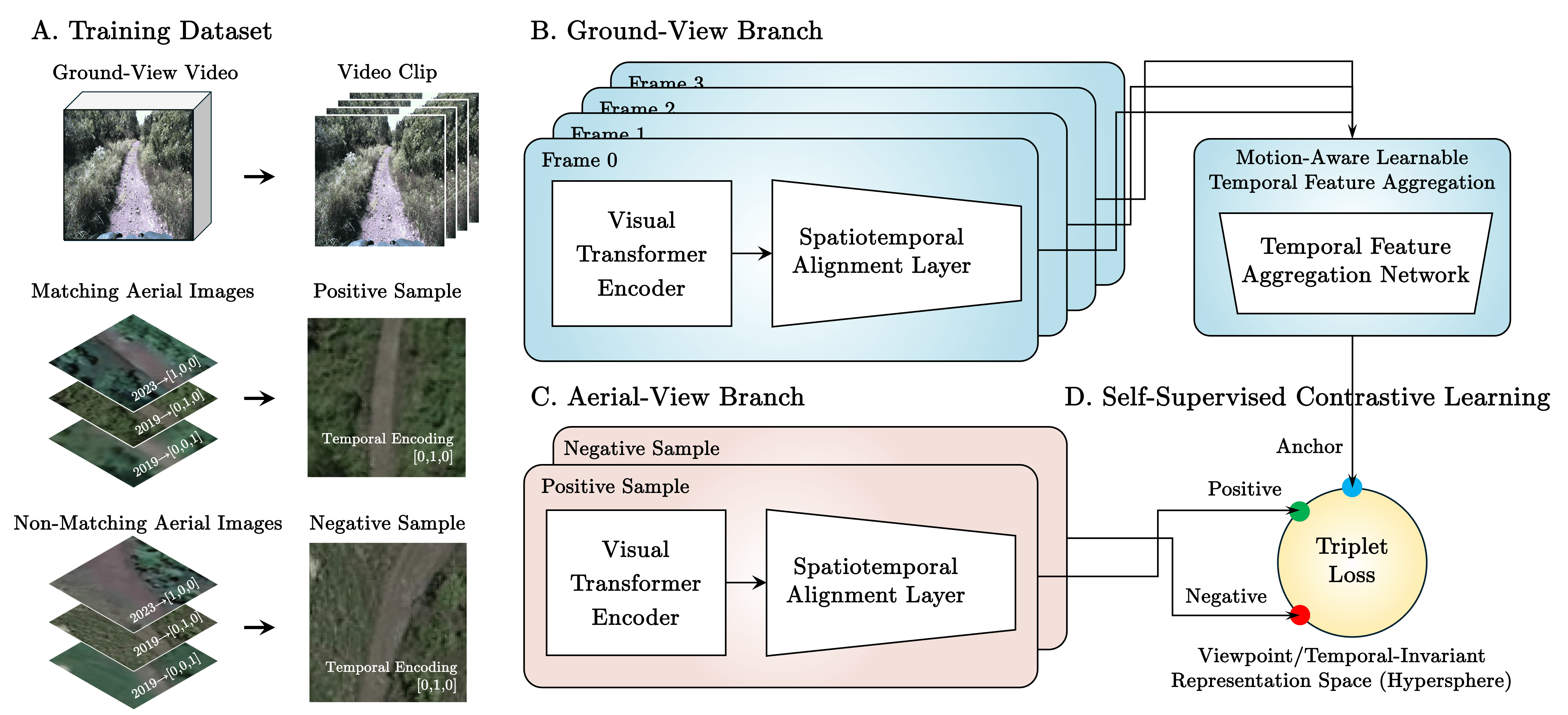}
    \vspace{-4mm}
    \caption{Network Architecture}
    \label{fig:network}
    \vspace{-6mm}
\end{figure*}

\section{Methodology}

\subsection{Spatiotemporal Contrastive Representations Learning} \label{method_section1}

\noindent We adopt a hierarchical training approach that first freezes the temporal aggregation module and trains the model on cross-view image pairs to learn viewpoint and temporal-invariant representations.

\noindent\textbf{Self-Supervised Contrastive Learning.}
Our objective is to learn a neural network $f_\Phi(\mathcal{G}, \mathcal{A}_k) \mapsto s \in [0, 1]$, parameterized by $\Phi$, which predicts a visual similarity score between a ground-view image $\mathcal{G}$ and a candidate aerial image $\mathcal{A}_k$. We interpret visual similarity as a proxy for geospatial proximity and orientation alignment.

We adopt a dual-branch architecture comprising two image encoders with identical structure: $f_g: \mathcal{G} \rightarrow \mathbb{R}^d$ for ground-view images and $f_a: \mathcal{A} \rightarrow \mathbb{R}^d$ for aerial images. Both branches use a frozen Vision Transformer (ViT) as the backbone, followed by a trainable spatiotemporal alignment layer that produces embedding vectors.
The encoders are trained using triplet loss, which formalizes our objective by comparing an anchor ground-view image $\mathcal{G}$ with a positive aerial image $\mathcal{A}_{\text{pos}}$ and a negative aerial image $\mathcal{A}_{\text{neg}}$:

{\small
\begin{equation}
\hspace*{-2.5em}
\mathcal{L}_{\text{triplet}} = \sum_{i=1}^{N} \left[ 
\left\| f_g(\mathcal{G}^i) - f_a(\mathcal{A}^i_{\text{pos}}) \right\|_2^2
- \left\| f_g(\mathcal{G}^i) - f_a(\mathcal{A}^i_{\text{neg}}) \right\|_2^2
+ \alpha \right]_+
\label{eq:triplet_loss}
\end{equation}
}

\noindent
where $\alpha$ is a margin hyperparameter and $[\cdot]_+ = \max(\cdot, 0)$ ensures the loss contributes only when the constraint is violated. We favor triplet loss for its stability and reduced computational complexity which enables us to train the model efficiently as demonstrated in the experiments. 

\noindent \textbf{Pose-Dependent Positive Sampling with Temporal Robustness Enforcement.} 
Unlike conventional approaches that discretize satellite imagery into uniform grid tiles and select the nearest tile as the positive sample ~\cite{vyas2022gama, downes2023wide, pillai2024garet, hausler2021patch}, we propose an unique positive sampling strategy that enforces both orientation sensitivity and temporal robustness.

For each ground-view image (used as the anchor) in training set, we sample multiple aerial patches from satellite images across different seasons, each centered at the robot's position and aligned to its heading. To prevent the model from exploiting superficial visual similarities, we employ hard positive mining—selecting the aerial sample whose embedding is furthest from the query embedding while still representing the same location. This forces the model to learn invariant features that persist despite significant appearance changes.

\noindent \textbf{Temporal-Matched Hard Negative Mining with Spatial and Orientation Hardness Control.}  
Prior works, such as \cite{vyas2022gama}, treat all aerial images in the batch that are not explicitly paired with the current ground-level query as negatives—regardless of their visual similarity or seasonal context. This indiscriminate treatment can result in trivial learning signals or false negatives, ultimately degrading model performance \cite{huynh2022boosting} as demonstrated in our experiments in Section \ref{exp:1}. 

To construct effective hard negatives without introducing false negatives, we implement a temporal-matched hard negative mining strategy constrained by both spatial proximity and orientation difference.
For each query ground-level image, we first select a hard positive aerial patch tagged with a temporal encoding $\tau$. We restrict all negative aerial patches to be sampled exclusively from the same temporal encoding as the positive. This eliminates the risk of the model learning to discriminate based on superficial seasonal differences rather than spatial and structural features. We demonstrated the importance of this temporal encoding through ablation study in Section \ref{exp:2}.

We impose spatial hardness by selecting negatives within a bounded distance from ground-truth and further apply an orientation constraint to encourage differentiation between the same location viewed from different directions:
\begin{equation}
d(\mathbf{x}_t, \mathbf{x}_j^-) \in [d_{\text{min}}, d_{\text{max}}] \quad \text{and} \quad |\theta_t - \theta_j^-| > \delta_\theta
\end{equation}

\subsection{Motion-Aware Learnable Temporal Feature Aggregation} \label{method_section2}

\noindent After image-pair pretraining, we freeze the image encoder and train the temporal aggregation module using video inputs to enhance the model’s robustness to visual outliers.

\noindent\textbf{Motion-Informed Video Frame Sampling.}
While prior works show that temporal aggregation improves pose estimation, most rely on uniformly sampled 0.5-second video clips~\cite{vyas2022gama} or fixed skip-frame selection~\cite{pillai2024garet}, which introduces redundant frames during slow motion and excessive spatial spread during fast motion, violating geometric consistency with the corresponding aerial image.

To address this, we propose a motion-informed frame sampling strategy that adaptively selects frames to ensure spatially uniform coverage of the recent trajectory. Since motion data is time-synchronized with the camera, we can backtrack from the current time and identify a trajectory segment that satisfies both a maximum duration and a minimum travel distance. Within this segment, we uniformly sample $N$ frames such that they are evenly spaced along the trajectory path, yielding a better geometric representation than time-uniform sampling.

Let the camera operate at a fixed frame rate $f$, and let $\{\mathbf{p}_\tau\}_{\tau = t - fT}^{t}$ denote the sequence of poses over the past $T$ seconds. We select $N$ frames $\{ I_{t_i} \}_{i=1}^N$ such that their corresponding poses $\{ \mathbf{p}_{t_i} \}$ satisfy:
\begin{equation}
\mathbf{p}_{t_i} \approx \text{interpolate}\left(\mathbf{p}_0 + \frac{i-1}{N-1} \cdot \mathbf{L} \right), \quad i = 1, \dots, N
\end{equation}
where $\mathbf{L}$ is the total displacement vector over the clip, and the interpolation finds the closest timestamp whose cumulative traveled distance matches the desired location. This effectively ensures:
\begin{equation}
\|\mathbf{p}_{t_{i+1}} - \mathbf{p}_{t_i}\| \approx \frac{1}{N-1} \cdot \|\mathbf{p}_{t_N} - \mathbf{p}_{t_1}\|
\end{equation}

\noindent\textbf{Quality-Aware Temporal Feature Aggregation.}  
Instead of using computationally expensive 3D CNNs or temporal transformers, we introduce a lightweight soft-attention module that predicts per-frame quality weights based on their relevance to the corresponding aerial image.

Given frame embeddings $\{\mathbf{e}_i\}_{i=1}^N \in \mathbb{R}^d$ and a reference aerial embedding $\mathbf{a} \in \mathbb{R}^d$, we compute cosine similarity for each frame:
\begin{equation}
s_i = \cos(\mathbf{e}_i, \mathbf{a}) = \frac{\mathbf{e}_i \cdot \mathbf{a}}{||\mathbf{e}_i|| \, ||\mathbf{a}||}
\label{eq:cosine_sim}
\end{equation}

Each similarity score $s_i$ is concatenated with its corresponding embedding $\mathbf{e}_i$, forming the input to a frame-level quality predictor:
\begin{equation}
o_i = \text{MLP}([\mathbf{e}_i; s_i]), \quad w_i = \frac{\exp(o_i / \beta)}{\sum_{j=1}^N \exp(o_j / \beta)}
\label{eq:softmax_temp}
\end{equation}
where $\beta$ is a temperature parameter that controls attention sharpness. The normalized weights $w_i$ are then used to compute the aggregated video embedding:
\begin{equation}
\mathbf{e}_{\text{agg}} = \sum_{i=1}^N w_i \, \mathbf{e}_i
\label{eq:weighted_agg}
\end{equation}

We train this module using a triplet loss adapted from Equation~\eqref{eq:triplet_loss}, with the aggregated video embedding $\mathbf{e}_{\text{agg}}$ replacing individual ground-frame embeddings:
\begin{equation}
\mathcal{L}_{\text{triplet}}^{\text{agg}} = \sum_{i=1}^{N} \left[ 
\left\| \mathbf{e}_{\text{agg}}^i - f_a(\mathcal{A}^i_{\text{pos}}) \right\|_2^2
- \left\| \mathbf{e}_{\text{agg}}^i - f_a(\mathcal{A}^i_{\text{neg}}) \right\|_2^2
+ \alpha \right]_+
\label{eq:agg_triplet_loss}
\end{equation}

To further stabilize training and promote diverse attention, we incorporate two regularization terms: a direct similarity term and an entropy penalty:
\begin{equation}
\mathcal{L} = \mathcal{L}_{\text{triplet}}^{\text{agg}} - \lambda_{\text{sim}} \cdot \cos(\mathbf{e}_{\text{agg}}, \mathbf{a}) - \lambda_{\text{H}} \cdot \mathcal{H}(w)
\label{eq:total_loss}
\end{equation}
where $\mathcal{H}(w) = -\sum_{i=1}^N w_i \log w_i$ is the entropy of the weight distribution, encouraging diversity by preventing the model from collapsing attention onto a single frame. The hyperparameters $\lambda_{\text{sim}}$ and $\lambda_{\text{H}}$ control the contributions of the similarity maximization and entropy regularization terms, respectively.

\subsection{Temporally-Consistent Data Augmentation} \label{method_section3}
To preserve temporal structure, prior work has introduced temporally consistent augmentations, where the same augmentation parameters are applied across all frames of a clip~\cite{qian2021spatiotemporal}. In this work, we extend this idea to the cross-view localization setting through a twofold augmentation strategy that ensures consistency both within and across modalities. 

\noindent\textbf{Intra-Clip Temporal Consistency.}  
We apply temporally consistent augmentations to all frames of the ground-level video clip $\{I_t\}_{t=1}^T$, using shared spatial (e.g., crop, resize) and appearance (e.g., brightness, contrast, hue) transformations. This preserves temporal coherence within the video and improves the model’s ability to learn consistent temporal features for downstream aggregation.

\noindent\textbf{Cross-View Consistency for Aerial Samples.}  
In our pose-dependent positive sampling strategy with temporal robustness enforcement, ground-view clips are paired with hard positive aerial images sampled from satellite imagery across seasons, as well as negative samples from the same temporal context. 
To prevent the model from exploiting spurious appearance cues introduced by augmentation differences, we apply the same augmentation parameters to each aerial-positive and its corresponding negatives. This enforces consistency across views and seasons, focusing learning on meaningful geometric and semantic alignment.

\subsection{Probabilistic Localization with Neural Perception and Adaptive Inference} \label{method_section4}

\noindent\textbf{Neural-Augmented Monte Carlo Localization.}  
Rather than relying on traditional handcrafted measurement models—often tuned to specific sensor modalities or assumptions about environmental regularity—we adopt a neural similarity function $f_\Phi(\cdot)$ as a learned measurement model within our particle filter. 
As shown in Section~\ref{exp:1}, the neural network maps a ground-level video clip and a candidate aerial patch to a similarity score that reflects geospatial alignment, demonstrating robustness to seasonal shifts, viewpoint changes, and illumination variation without requiring environment-specific tuning.

\noindent\textbf{Uncertainty-Aware Likelihood Modulation via Spatial KDE.}  
While adaptive temperature tuning has been explored in other domains (e.g., attention scaling or confidence-aware filtering) \cite{fox2001kld}, it is rarely grounded in spatial belief uncertainty within a neural-augmented Monte Carlo framework. 
To prevent filter collapse in visually ambiguous regions,  we introduce an uncertainty quantification technique based on the entropy of the spatial particle distribution. We estimate the density over particle positions $(x, y)$ using kernel density estimation (KDE):

\begin{equation}
p_t(\mathbf{x}) = \sum_{n=1}^{M} w_t^{[n]} K_h\left( \mathbf{x} - (x_t^{[n]}, y_t^{[n]}) \right)
\end{equation}

\noindent where $K_h$ is a Gaussian kernel with bandwidth $h$. The entropy $H_t$ of this density captures the spatial uncertainty in the current belief:

\begin{equation}
H_t = -\int p_t(\mathbf{x}) \log p_t(\mathbf{x}) \, d\mathbf{x}
\end{equation}

This entropy is then used to adaptively modulate the softmax temperature $\lambda_t$ in the observation model:

\begin{equation}
\lambda_t = \lambda_{\text{base}} \cdot \exp(-\gamma \cdot H_t)
\end{equation}

\section{Experiments}

\subsection{Implementation Details}\label{exp:inplementation details}

\begin{figure}
    \centering
    \includegraphics[width=1.0\linewidth]{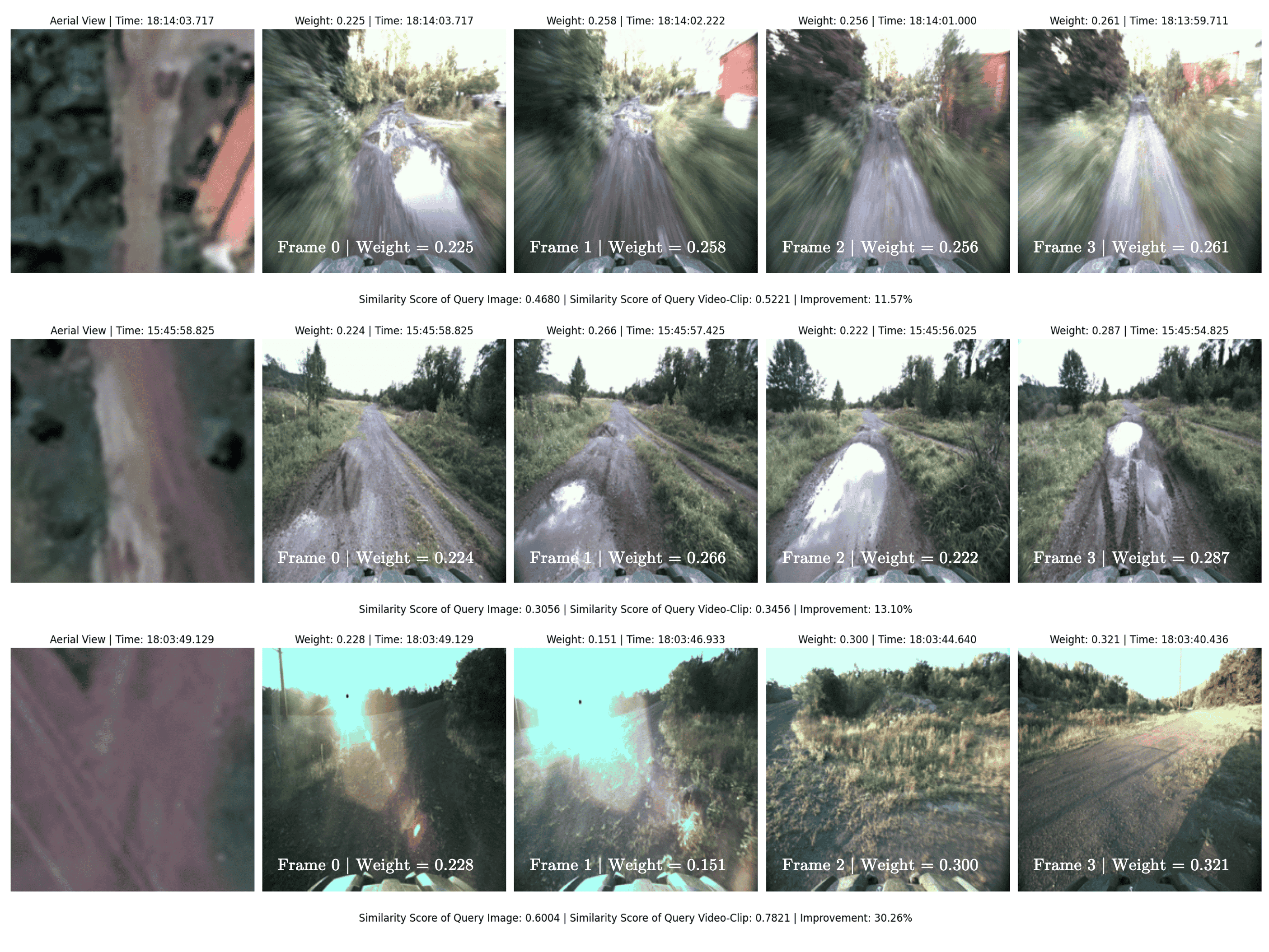}
    \vspace{-8mm}
    \caption{Visualization of frame quality weights from the temporal aggregation module. The model assigns lower weights to visually unreliable frames affected by transient occlusions or perceptual aliasing (e.g., water puddles, lens flare), and emphasizes more informative observations. This demonstrates greater robustness over single-frame image matching.}
    \label{fig:when-attention}
    \vspace{-6mm}
\end{figure}

\noindent \textbf{Network Architecture.}  
MoViX uses a dual-branch architecture with frozen ViT-B/16 backbones and trainable alignment layers $[768, 1024, 512]$ to produce $L_2$-normalized 512-dimensional embeddings. 
The temporal aggregation MLP takes each frame's 512-D embedding concatenated with its cosine similarity as input (513-D total) and outputs a frame quality score through hidden sizes $[513, 256, 128, 1]$ with ReLU activations. Softmax-normalized scores serve as per-frame weights for weighted pooling, emphasizing informative frames during cross-view matching. See Fig.~\ref{fig:when-attention} for an example of how weights are distributed across frames in a video clip.

\vspace{0.5em}
\noindent \textbf{Data Preprocessing and Augmentation.}
We construct ground-aerial training pairs in a self-supervised manner using time-synchronized camera and odometry data, aligned via a rigid transform from UTM to the map frame. Positive pairs are selected based on pose, leveraging historical satellite imagery to encourage the learning of temporally invariant features under long-term environmental changes.
Hard negatives are sampled within a 5--40 meter radius and filtered by a heading difference constraint ($\delta_\theta \geq 30^\circ$). Video clips are generated using motion-informed sampling to ensure spatially uniform frame selection, with a maximum spacing of 5 meters along the trajectory. Aerial patches are cropped to 20$\times$20 meters to approximate the ground-view field of view.

\vspace{0.5em}
\noindent \textbf{Training Protocol.}  
Training proceeds in two stages. First, we train the cross-view embedding network with a triplet loss ($\alpha = 0.2$) using the Adam optimizer (learning rate = 0.001) and gradient clipping (norm = 1.0). A reduce-on-plateau scheduler lowers the learning rate by 0.5 after 3 stagnant epochs, with early stopping applied over a 40-epoch budget. Next, we freeze the image encoders and train the temporal aggregation module for 30 epochs using the same optimizer. The loss combines triplet loss with direct similarity maximization ($\lambda_{\text{sim}} = 0.1$) and entropy regularization ($\lambda_{\text{H}} = 0.1$).

\begin{figure}
    \centering
    \includegraphics[width=1\linewidth]{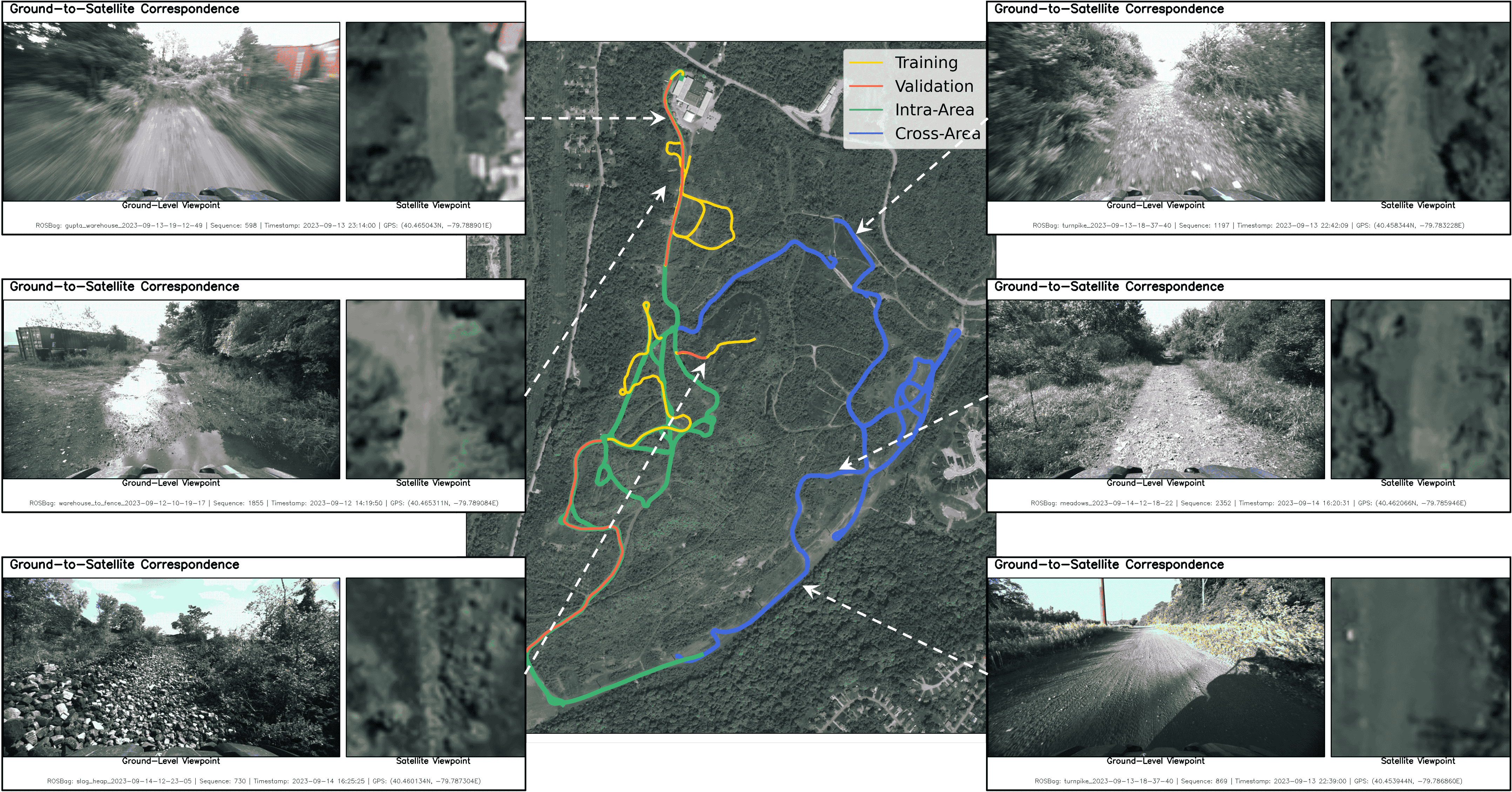}
    \vspace{-6mm}
    \caption{
    Training, Validation, and Test Split from the TartanDrive 2.0 dataset. The training set includes 5 intra-area trajectories totaling 2.91 km (3,174 keyframes), corresponding to less than 30 minutes of driving data. Validation is performed on 3 intra-area trajectories spanning 1.37 km (1,405 keyframes). The test set consists of 8 trajectories covering 12.29 km in total, comprising 4 intra-area (8.08 km) and 4 cross-area (4.21 km) segments.
    }
    \label{fig:data-split}
    \vspace{-6mm}
\end{figure}

\vspace{0.5em}
\noindent \textbf{Training and Evaluation Data.}  
We conduct our experiments using the TartanDrive 2.0 dataset, collected in Pittsburgh, PA, between September and November 2023 using a Yamaha Viking vehicle. Our model is trained on less than 30 minutes of driving data, demonstrating the efficiency and data-effectiveness of our approach.  An overview of the training, validation, and test split is shown in Fig.~\ref{fig:data-split}.
Satellite imagery is obtained via the Google Maps Static API~\cite{googlemapsapi}. For training and validation, we use summer imagery from May 2023, September 2019, and August 2016. To assess temporal robustness, we reserve fall imagery—captured in November 2021, October 2020, and October 2018—exclusively for testing.

To further evaluate our model’s generalization to out-of-distribution environments, we collected an additional real-world off-road dataset in August 2024 in Baltimore, MD, using a ClearPath Warthog robot. This test set includes satellite imagery from November 2024 and June 2022. Since our method does not require camera calibration and is resilient to outdated satellite imagery, it enables robust deployment across platforms and environments, regardless of camera height, tilt angle, or seasonal mismatch in the reference map.

\vspace{0.5em}
\noindent \textbf{Inference Pipeline.}  
At test time, our cross-view neural matching module serves as a measurement model within an MCL framework with 300 particles. Particles are initialized with Gaussian noise around the prior pose and propagated using noisy odometry, injecting additional translation and rotation noise.  At every timestep, satellite patches centered on each particle are cropped and compared to the ground-view input using the learned similarity function. Our approach does not require any environment-specific parameter tuning; the same set of parameters is used consistently across all test trials.

\begin{figure}[t]
    \centering
    \includegraphics[width=0.95\linewidth]{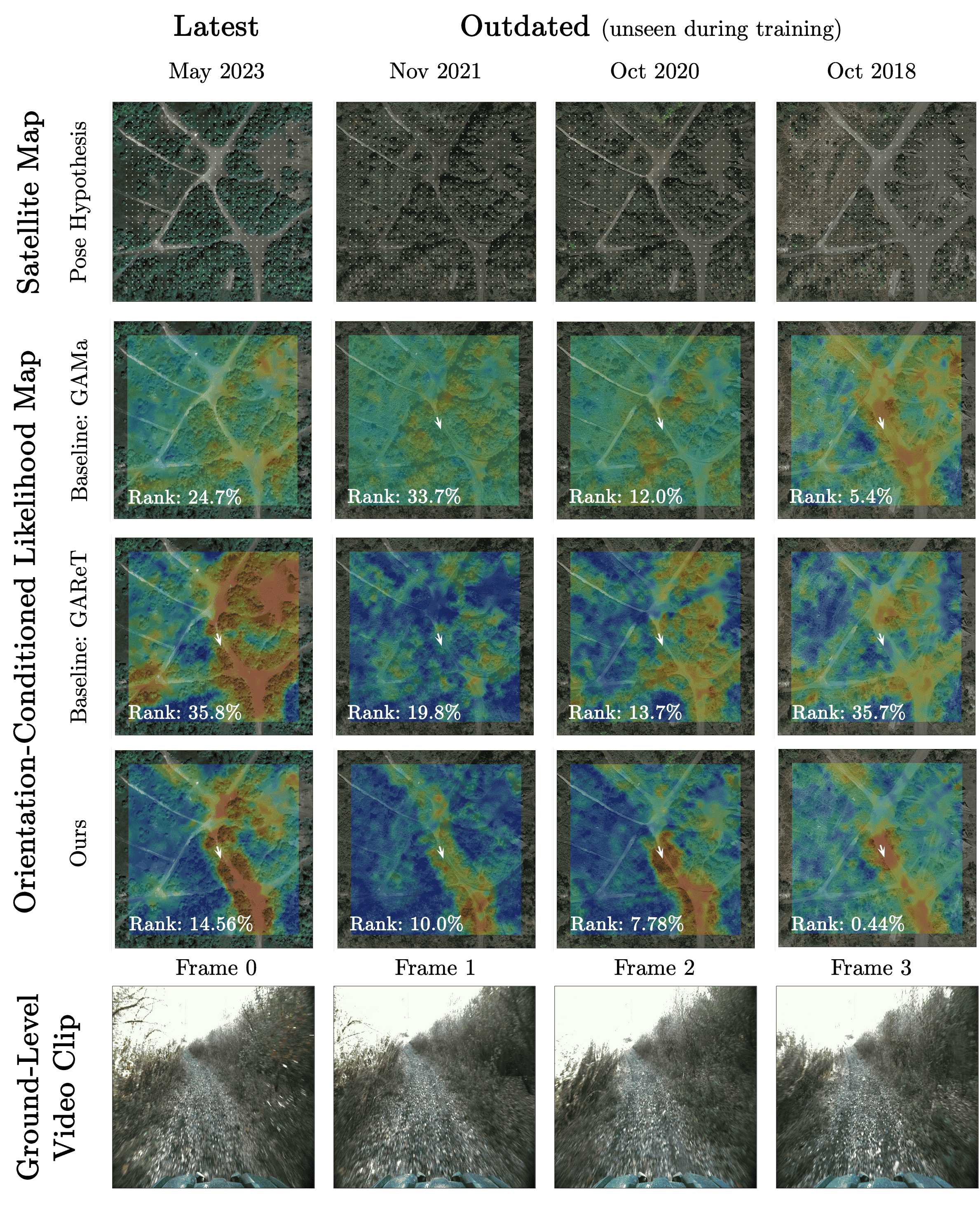}
    \vspace{-3mm}
    \caption{An orientation-conditioned likelihood map that approximates $P(\text{observation} \mid \text{pose}, \text{map})$. The $30 \times 30$ spatial grid defines candidate $(x, y)$ locations, with the orientation $\theta$ fixed at $-67.5\degree$. For each grid cell, an aerial patch aligned to the orientation prior is compared with the ground-level observation to compute a similarity score, normalized to [0,1] (red: high, green: medium, blue: low) and interpolated for visualization. The white arrow indicates the ground-truth pose. Localization accuracy is evaluated by ranking the ground-truth pose among all 900 candidates (lower percentile is better). This region is unseen during training.
    }
    \vspace{-6mm}
    \label{fig:pose-estimation}
\end{figure}

\subsection{Frame-Level 3-DoF Pose Estimation}\label{exp:1}

We evaluate our cross-view neural matching module’s ability to localize a ground-level observation within a $150 \times 150$ meter satellite map centered on the ground-truth location, using a rough camera orientation as a prior. The map is discretized into a $30 \times 30$ grid, and at each location, a $20 \times 20$ meter aerial patch aligned to the orientation prior is extracted as a candidate. Each candidate is compared against the ground-level observation using the learned similarity function to produce a matching score. These scores approximate the observation likelihood $P(\text{observation} \mid \text{pose}, \text{map})$. Fig.~\ref{fig:pose-estimation} presents a qualitative comparison of the resulting orientation-conditioned likelihood maps generated by different models.

\noindent \textbf{Baseline Analysis.}
GAMa~\cite{vyas2022gama} adopts NT-Xent loss~\cite{sohn2016improved}, an InfoNCE-based contrastive loss that treats all other samples in a batch as negatives. While this avoids explicit negative mining, it risks introducing false negatives—particularly in natural environments where many distinct locations along the vehicle's path share similar visual features (e.g., vegetation-lined corridors). As a result, semantically correct matches may be penalized, weakening the learning signal.
GAReT~\cite{pillai2024garet} mitigates this by using a triplet loss, but selects positives based only on proximity in a discretized satellite grid, without considering orientation. This leads to poor directional sensitivity, as visually similar yet misaligned patches may still be treated as correct.

\begin{figure}[t]
    \centering
    \includegraphics[width=0.65\linewidth]{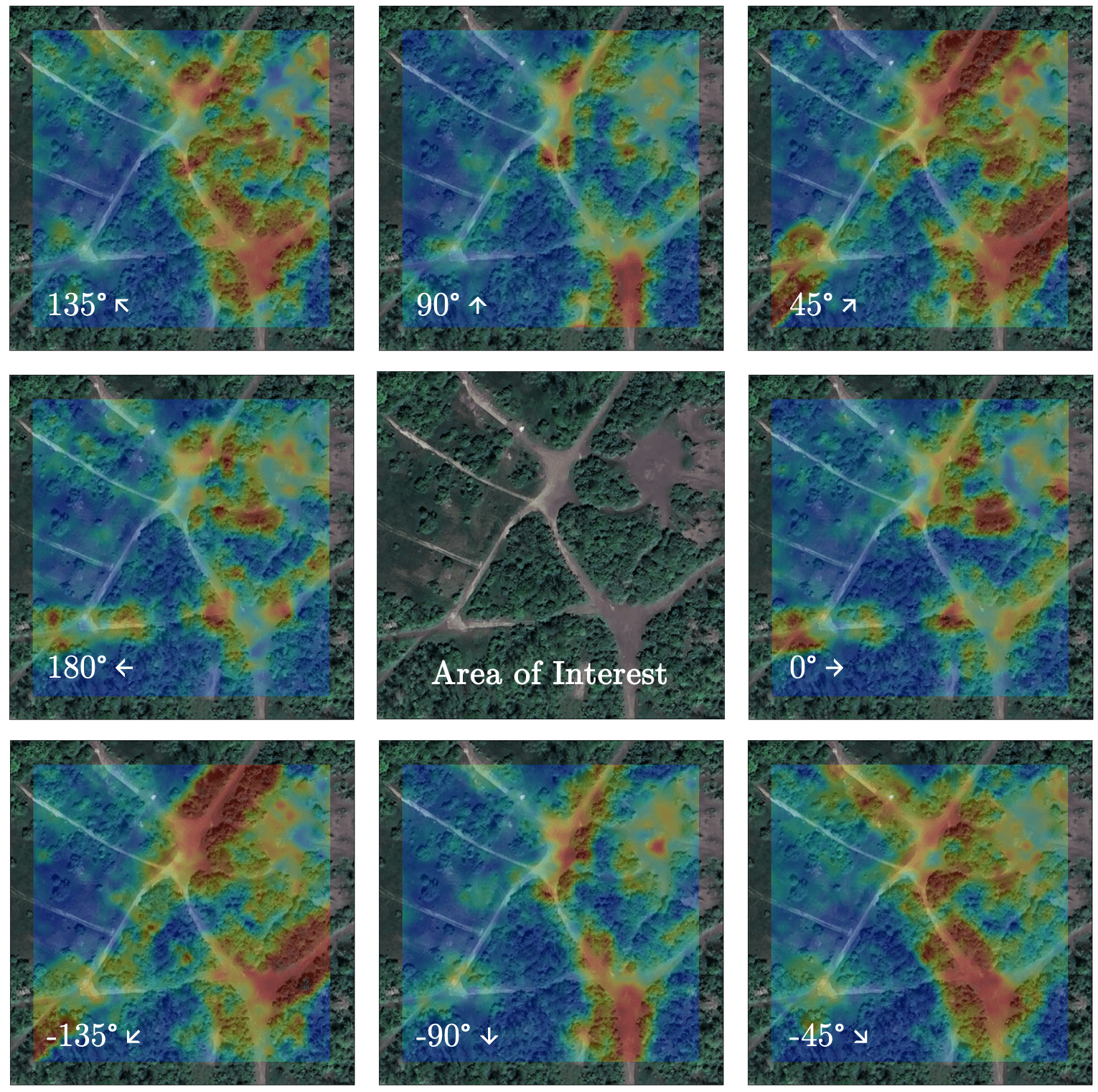}
    \vspace{-3mm}
    \caption{
    Orientation-conditioned likelihood maps generated by our method under varying heading priors ($\theta \in (-180^\circ, 180^\circ]$, in $45^\circ$ increments), using the same ground-level observation as in Fig.~\ref{fig:pose-estimation} and satellite imagery from May 2023. The likelihood distribution shifts consistently with the heading prior, highlighting the model’s directional sensitivity and its ability to disambiguate pose hypotheses based on orientation cues.
    }
    \label{fig:orientation-prior}
    \vspace{-6mm}
\end{figure}

In contrast, our method explicitly conditions positive pair selection on both spatial location and orientation, reducing false negatives and improving directional alignment during training. While baseline methods degrade under distribution shifts caused by outdated satellite imagery, our model is trained with multi-season overhead data and learns temporally invariant cross-view features. This enables the generation of sharper, orientation-aware likelihood maps that preserve multiple plausible hypotheses under perceptual ambiguity. As shown in Fig.~\ref{fig:orientation-prior}, our model responds coherently to changes in the orientation prior, demonstrating strong directional sensitivity and the ability to resolve pose ambiguities based on orientation.

\begin{table*}[t]
\caption{Comparison of Trajectory Estimation with Season-Matched Satellite Imagery}
\vspace{-0.3cm}
\label{tab:same-season}
\footnotesize
\begin{center}
\begin{tabular}{l|cccccc|cccccc}
\toprule
\multirow{3}{*}{\textbf{Method}} & \multicolumn{6}{c|}{\textbf{Intra-Area (Same-Season)}} & \multicolumn{6}{c}{\textbf{Cross-Area (Same-Season)}} \\
\cmidrule(lr){2-7} \cmidrule(lr){8-13}
& \multicolumn{2}{c}{$\downarrow$ ATE} & $\downarrow$ SDR & \multicolumn{3}{c|}{$\uparrow$ SR (\%)} & \multicolumn{2}{c}{$\downarrow$ ATE} & $\downarrow$ SDR & \multicolumn{3}{c}{$\uparrow$ SR (\%)} \\
& Mean & Max & (\%) & @10 & @25 & @50 & Mean & Max & (\%) & @10 & @25 & @50 \\
\midrule
Vidual Odom. & 210.6 & 449.0 & 27.4 & 4.0 & 5.4 & 7.9 & 98.8 & 213.1 & 18.0 & 14.4 & 27.3 & 41.7 \\
Lidar Odom. & 16.3 & 37.4 & 1.1 & 29.9 & 85.7 & 98.3 & 29.2 & 66.4 & 5.6 & 75.8 & 77.3 & 80.3 \\
Lidar Odom. (NF) & 14.0 & 32.0 & 1.3 & 39.5 & 88.1 & 100.0 & 3.0 & 9.1 & 0.7 & 99.7 & 100.0 & 100.0 \\
\midrule
ViT & 106.3 & 266.3 & 15.7 & 13.2 & 28.8 & 40.1 & 72.3 & 172.5 & 15.6 & 27.7 & 60.8 & 68.5 \\
DINOv2 & 60.0 & 150.4 & 16.5 & 16.0 & 44.2 & 65.0 & 72.0 & 174.3 & 18.4 & 14.2 & 35.4 & 58.8 \\
GAMa & 28.3 & 98.7 & 11.9 & 36.3 & 65.8 & 83.5 & 29.2 & 82.2 & 14.6 & 40.9 & 63.7 & 73.2 \\
GAReT & 7.2 & 19.6 & 2.7 & 72.2 & 100.0 & 100.0 & 11.2 & 31.5 & 6.0 & 58.5 & 94.0 & 99.7 \\
Ours ($-$TRE) & 8.3 & 28.8 & 3.2 & 69.5 & 96.6 & 100.0 & 11.8 & 39.8 & 6.0 & 68.4 & 91.3 & 95.4 \\
Ours ($-$TRE $+$ CJ) & 8.7 & 27.4 & 2.8 & 70.3 & 96.1 & 100.0 & 9.3 & 25.4 & 5.3 & 63.7 & 98.5 & 100.0 \\
\rowcolor{gray!15}
Ours & 8.3 & 22.5 & 2.2 & 68.2 & 99.0 & 100.0 & 9.8 & 27.4 & 6.6 & 59.7 & 96.8 & 100.0 \\
\bottomrule
\end{tabular}
\end{center}

\vspace{-0.4cm}
\end{table*}

\begin{table*}[t]
\caption{Comparison of Trajectory Estimation with Outdated Satellite Imagegy}
\vspace{-0.3cm}
\label{tab:cross-season}
\footnotesize
\begin{center}
\begin{tabular}{l|cccccc|cccccc}
\toprule
\multirow{3}{*}{\textbf{Method}} & \multicolumn{6}{c|}{\textbf{Intra-Area (Cross-Season)}} & \multicolumn{6}{c}{\textbf{Cross-Area (Cross-Season)}} \\
\cmidrule(lr){2-7} \cmidrule(lr){8-13}
& \multicolumn{2}{c}{$\downarrow$ ATE} & $\downarrow$ SDR & \multicolumn{3}{c|}{$\uparrow$ SR (\%)} & \multicolumn{2}{c}{$\downarrow$ ATE} & $\downarrow$ SDR & \multicolumn{3}{c}{$\uparrow$ SR (\%)} \\
& Mean & Max & (\%) & @10 & @25 & @50 & Mean & Max & (\%) & @10 & @25 & @50 \\
\midrule
ViT & 85.6 & 237.0 & 18.6 & 12.8 & 35.1 & 56.8 & 89.0 & 236.7 & 22.6 & 16.7 & 30.8 & 45.8 \\
DINOv2 & 87.7 & 243.6 & 13.9 & 13.8 & 35.5 & 56.0 & 77.0 & 177.3 & 18.8 & 11.6 & 27.9 & 44.6 \\
GAMa & 120.0 & 305.6 & 27.8 & 13.6 & 26.7 & 45.5 & 34.5 & 87.4 & 18.2 & 24.7 & 55.2 & 78.2 \\
GAReT & 22.8 & 72.8 & 4.6 & 59.0 & 89.3 & 93.4 & 26.1 & 61.9 & 9.9 & 37.8 & 74.7 & 91.4 \\
Ours ($-$TRE) & 15.7 & 58.0 & 4.7 & 60.6 & 88.7 & 95.7 & 22.3 & 63.6 & 9.7 & 41.9 & 74.7 & 88.0 \\
Ours ($-$TRE $+$ CJ) & 9.2 & 26.4 & 2.6 & 67.5 & 95.8 & 100.0 & 23.3 & 65.0 & 7.4 & 53.7 & 85.6 & 94.7 \\
Ours ($-$TE) & 9.2 & 28.2 & 2.8 & 68.2 & 95.9 & 100.0 & 17.7 & 47.4 & 7.1 & 52.8 & 81.2 & 91.0 \\
\rowcolor{gray!15}
Ours & 7.9 & 23.4 & 2.6 & 74.6 & 98.3 & 100.0 & 10.7 & 30.0 & 6.9 & 58.9 & 93.7 & 100.0 \\
\bottomrule
\end{tabular}
\end{center}
\vspace{-0.3cm}
\begin{tablenotes}
\scriptsize
\item 

\end{tablenotes}
\vspace{-0.6cm}
\end{table*}

\subsection{Probabilistic 3-DoF Trajectory Estimation}\label{exp:2}

We evaluate MoViX on continuous 3-DoF trajectory estimation from ground-level video inputs across both intra-area and cross-area settings, using either same-season or cross-season satellite imagery as a global reference map.

\noindent \textbf{Evaluation Metrics.}  
Absolute Trajectory Error (ATE) measures the average Euclidean distance (in meters) between predicted and ground-truth positions over time, capturing the overall positional accuracy of the trajectory.
Scale Drift Rate (SDR) quantifies the relative error in trajectory length by comparing the total distance of the predicted path to that of the ground truth, providing insight into accumulated scale drift.
Success Rate (SR) evaluates localization accuracy by computing the percentage of frames where the predicted position falls within a specified distance threshold of the ground truth. We report SR at 10m, 25m, and 50m thresholds.

\noindent \textbf{Baselines.}  
We compare MoViX against two categories of methods. (1) Odometry-based approaches: TartanVO~\cite{wang2021tartanvo} (visual odometry) and SuperOdometry~\cite{zhao2021super} (LiDAR-visual-inertial fusion), evaluated only in same-season settings since they do not use satellite imagery. (2) Cross-view localization methods: including GAMa~\cite{vyas2022gama}, GAReT~\cite{pillai2024garet}, and vision backbones like ViT~\cite{dosovitskiy2020image} and DINOv2~\cite{oquab2023dinov2}. For fair comparison, all models are integrated into our particle filtering framework (Section~\ref{method_section4}) using identical parameters (Section~\ref{exp:inplementation details}) to estimate trajectories.

\noindent \textbf{Ablation Studies.}  
To understand the contribution of each component in our framework, we evaluate ablated variants of our model, particularly in settings with outdated satellite imagery where baseline methods typically fail to generalize.
(1) \textit{Ours $-$TRE} removes Temporal Robustness Enforcement (TRE) by training only on single-season satellite imagery, though it retains pose-dependent positive sampling. (2) \textit{Ours $-$TRE $+$ CJ} explores whether strong appearance augmentation can substitute for temporal diversity. It applies Color Jetting (CJ) to single-season imagery during training to simulate visual variability. (3) \textit{Ours $-$TE} uses multi-season satellite imagery but omits Temporal Encoding (TE) from the contrastive loss, disabling the model’s ability to explicitly learn temporal invariance.

\noindent \textbf{Evaluation with Season-Matched Satellite Imagery.}  
We evaluate all methods using satellite imagery from May 2023, which matches the season of the ground-level observations (see Table~\ref{tab:same-season}).

Odometry-based methods exhibit expected behavior: monocular visual odometry (TartanVO) accumulates significant drift due to the absence of global correction, while SuperOdometry (LiDAR-visual-inertial fusion) performs reliably in most cases but still struggles in some trajectories with sparse geometric features. We also report a variant, SuperOdometry (NF), which excludes failure segments for fair comparison—its performance reflects the upper bound for LiDAR-based odometry under ideal conditions.
Cross-view models with ViT or DINOv2 backbones struggle to bridge the large viewpoint and modality gap between ground and aerial imagery, resulting in mean ATE values exceeding 60 meters. GAMa improves on this by employing contrastive learning, but its sampling strategy is sensitive to false negatives, especially in visually ambiguous scenes with repeated textures. GAReT performs well in seen environments (achieving 100\% SR@25m in intra-area), but its lack of orientation awareness hinders generalization to unseen areas.

In contrast, our method, along with its ablated variants, shows strong robustness across both intra-area and cross-area settings. Its orientation-aware matching strategy enables more consistent localization, even under viewpoint ambiguities, highlighting the effectiveness of explicitly modeling directional cues.

\noindent \textbf{Evaluation with Outdated Satellite Imagery.}  
To evaluate robustness to temporal shifts, we focus on ablation variants of our method, as baseline models trained solely on May 2023 satellite imagery fail to generalize to outdated reference maps due to the lack of temporal resilience.

\textit{Ours $-$TRE} and \textit{Ours $-$TRE $+$ CJ} remove Temporal Robustness Enforcement (TRE), training only on May 2023 satellite imagery to match the baselines’ temporal setting. (1) \textit{Ours $-$TRE} retains pose-dependent positive sampling, improving ATE via better orientation alignment, but still fails to generalize to unseen seasons—highlighting the limits of single-season training. (2) \textit{Ours $-$TRE $+$ CJ} tests whether aggressive appearance augmentation (Color Jetting) can compensate for temporal gaps. While it achieves 100\% SR@50m in intra-area tests, its performance drops in cross-area settings, showing that superficial color shifts fail to capture deeper structural variations. (3) \textit{Ours $-$TE} includes multi-season imagery but omits Temporal Encoding (TE), leading to better ATE but lower success rates, as it often overfits to seasonal artifacts rather than learning consistent spatial features—emphasizing the importance of temporal alignment in contrastive learning.

In contrast, our full MoViX model combines multi-season training, orientation-aware supervision, and temporally robust encoding. This design consistently achieves the best localization performance across both seasonal and geographic shifts. 
We showcase the application of \textit{MoViX} on our off-road Baltimore dataset—an out-of-distribution scenario collected in a geographically distinct environment with a different robot platform—in the accompanying video (\href{https://youtu.be/y5wL8nUEuH0}{https://youtu.be/y5wL8nUEuH0}). This demonstration highlights the model’s robustness to platform variation, geographic shift, and temporal mismatch in satellite imagery.

\section{Conclusion} 
\label{sec:conclusion}
We presented MoViX, a self-supervised framework for robust cross-view video localization in GPS-denied, off-road environments. By learning representations that are robust to viewpoint and seasonal variation, while preserving directional cues critical for localization, MoViX demonstrates strong generalization across spatial and temporal variations. A key limitation is its reliance on GPS-fused odometry during training to establish ground-satellite correspondences, which may limit applicability in regions where such priors are unavailable. Future work may explore ways to relax this dependency, moving toward broader deployment in fully infrastructure-free environments.

{\fontsize{6.5pt}{7.5pt}\selectfont
\bibliographystyle{unsrt}
\bibliography{references}
}

\end{document}